\documentclass{article}

\usepackage[preprint, nonatbib]{neurips_2024}

\usepackage[utf8]{inputenc} 
\usepackage[T1]{fontenc}    
\usepackage{hyperref}       
\usepackage{url}            
\usepackage{booktabs}       
\usepackage{amsfonts}       
\usepackage{nicefrac}       
\usepackage{microtype}      
\usepackage{newunicodechar}
\usepackage[utf8]{inputenc} 
\usepackage[table,xcdraw]{xcolor}
\usepackage{paralist}
\usepackage{todonotes}
\usepackage{CJKutf8}
\usepackage{authblk}
\usepackage{graphicx}
\usepackage{subcaption}
\usepackage{float}
\usepackage{wrapfig}
\usepackage[labelfont=bf,labelsep=period]{caption}
\usepackage{colortbl}
\usepackage{multirow}
\usepackage{makecell} 
\usepackage{array}
\usepackage{arydshln} 
\usepackage{booktabs} 
\usepackage{geometry}
\usepackage{array}
\usepackage{xcolor}
\usepackage{tikz}
\usepackage{arydshln} 
\usepackage{mdframed}
\newmdenv[innerleftmargin=-28pt,innerrightmargin=-28pt,skipabove=10pt,skipbelow=10pt,linecolor=black,linewidth=0.5pt]{quoteBox}

\usepackage{listings}       
\usepackage{enumitem} 
\setlist[enumerate]{  
  leftmargin=16pt,      
  itemindent=1pt,    
  labelsep=5pt,       
  topsep=0pt,         
  partopsep=0pt,      
  parsep=2pt,         
  itemsep=2pt         
}

\setlist[itemize]{  
  leftmargin=16pt,      
  itemindent=1pt,    
  labelsep=5pt,       
  topsep=0pt,         
  partopsep=0pt,      
  parsep=2pt,         
  itemsep=2pt         
}

\title{Routine: A Structural Planning Framework for LLM Agent System in Enterprise}

\author{Guancheng Zeng\thanks{Equal contribution.}  $^{\ 1}$}
\author{Xueyi Chen$^{*}$}
\author{Jiawang Hu$^{*}$}
\author{Shaohua Qi$^{*}$}
\author{Yaxuan Mao}
\author{Zhantao Wang}
\author{Yifan Nie}
\author{Shuang Li}
\author{Qiuyang Feng}
\author{Pengxu Qiu}
\author{Yujia Wang}
\author{Wenqiang Han}
\author{Linyan Huang}
\author{Gang Li}
\author{Jingjing Mo}
\author{Haowen Hu\thanks{Corresponding Author}  $^{\ 2}$}
\affil{Digital China AI Research \\ 
\{cenggc\textsuperscript{1}, huhwa\textsuperscript{2}\}@digitalchina.com}

\begin{document}

\maketitle

\begin{abstract}
The deployment of agent systems in an enterprise environment is often hindered by several challenges: common models lack domain-specific process knowledge, leading to disorganized plans, missing key tools, and poor execution stability. To address this, this paper introduces \textbf{Routine}, a multi-step agent planning framework designed with a clear structure, explicit instructions, and seamless parameter passing to guide the agent's execution module in performing multi-step tool-calling tasks with high stability. In evaluations conducted within a real-world enterprise scenario, Routine significantly increases the execution accuracy in model tool calls, increasing the performance of \textbf{GPT-4o} from 41.1\% to \textbf{96.3\%}, and \textbf{Qwen3-14B} from 32.6\% to \textbf{83.3\%}. We further constructed a Routine-following training dataset and fine-tuned Qwen3-14B, resulting in an accuracy increase to \textbf{88.2\%} on scenario-specific evaluations, indicating improved adherence to execution plans. In addition, we employed Routine-based distillation to create a scenario-specific, multi-step tool-calling dataset. Fine-tuning on this distilled dataset raised the model's accuracy to \textbf{95.5\%}, approaching GPT-4o's performance. These results highlight Routine's effectiveness in distilling domain-specific tool-usage patterns and enhancing model adaptability to new scenarios. Our experimental results demonstrate that Routine provides a practical and accessible approach to building stable agent workflows, accelerating the deployment and adoption of agent systems in enterprise environments, and advancing the technical vision of AI for Process.


\end{abstract}

\section{Introduction}

The emergence of autonomous agents has given rise to the active development of various frameworks and functional modules in industry, bringing diverse structures and capabilities to the field~\cite{wooldridge2009introduction}. Since both research and real-world practice have progressed, agent architectures have gradually converged towards more stable designs. A typical agent system consists of four key modules: planning, execution, tools, and memory; they collaborate to accomplish complex tasks and maintain coherent interactions with users~\cite{Wang_2024}. These modules are typically driven by large language models (LLMs), which enable agents to adapt to common scenarios and perform tasks such as data analysis, report summarization, and web interface manipulation with strong generalization and generative capabilities~\cite{hong2024datainterpreterllmagent, patil2023gorilla, dasilva2025extractingknowledgegraphsuser}. Numerous high-performing and innovative agent applications have emerged, demonstrating the potential of LLM-based agent systems in practical use~\cite{shen2025mindmachinerisemanus}.

However, in enterprise-specific scenarios, it is still hard to implement agent systems due to their inability to integrate reliably with scenario-specific tools and effectively solve real-world tasks~\cite{kandogan2025orchestratingagentsdataenterprise}. Common planning models often fail to generate robust execution plans and tend to overlook essential tools -- particularly those involving permission verification and model generation -- due to the lack of scenario-specific knowledge~\cite{xiao2024flowbenchrevisitingbenchmarkingworkflowguided}. This issue is further compounded by insufficient tool descriptions in real-world scenarios, making it difficult for models to select the appropriate tools and parameters, leading to instability in task execution~\cite{xu2025reducingtoolhallucinationreliability}.
Additionally, current agent planning lacks a unified, structured, and complete format~\cite{ xiao2024flowbenchrevisitingbenchmarkingworkflowguided, zeng2024flowmindautomaticworkflowgeneration, sonkin2025beyond}. As a result, agents often passed in custom formats to the execution module, which must infer tool-calling instructions without clear structural guidance~\cite{minkova2024wordsworkflowsautomatingbusiness}.
This ambiguity leads to mismatch between planning steps and tool calls. 
While low-code platforms offer a more stable alternative, they still depend heavily on manual effort, limiting the efficiency of workflow development~\cite{wornow2024automatingenterprisefoundationmodels, shlomov2024idabreakingbarriersnocode}.

\begin{wrapfigure}{r}{0.55\linewidth} 
    \vspace{-0.1em}
    \centering
    \includegraphics[width=1\linewidth]{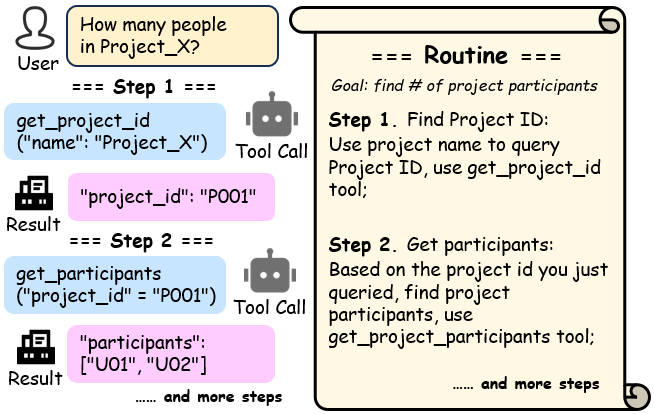}
    \vspace{-1.6em}
    \caption{The mechanism of Routine guiding an LLM agent through multi-step tool calls}
    \label{fig: routineIntro}
\end{wrapfigure}

In this paper, we propose \textbf{Routine}~\cite{openai_cookbook_using_reasoning_2024}, a structured planning framework for agents to address the above issues; the mechanism is shown in Figure \ref{fig: routineIntro}~\cite{xiao2024flowbenchrevisitingbenchmarkingworkflowguided, liu2025toolplannertaskplanningclusters}. Given a planning prompt from an expert, the planning model refines details of the procedure and transforms them into a well-formatted natural language Routine, which is then passed to the execution model to perform multi-step tool calling and solve user problems. In experiments, GPT-4o achieved an overall accuracy of 96.3\% on the scenario-specific test set with Routine planning, significantly outperforming its accuracy of 41.1\% without Routine. Similarly, Qwen3-14B's performance improved from 32.6\% to 83.3\% with the incorporation of Routine. 

Furthermore, we synthesized a total of 4,209 common training data with Routine generation based on an open multi-tool dataset\footnote{https://github.com/PKU-Baichuan-MLSystemLab/BUTTON}. After training, Qwen3-14B achieved an accuracy of 88.2\%, further improving its Routine instruction-following capability. Additionally, we performed data distillation based on Routine in real enterprise scenarios, generating 537 scenario-specific, single-turn, multi-step tool calling samples. Qwen3-14B trained on this distilled dataset achieved an accuracy of 95.5\% on the test set, approaching GPT-4o's 96.3\%, demonstrating the potential of using Routine-based distillation to adapt execution models to a specific process.

In summary, our contributions include: 
\begin{enumerate}
\item We propose Routine, a well-formatted planning framework that improves the stability of multi-step tool calling in agent systems within enterprise scenarios;
\item We generate a Routine-compliant training dataset and fine-tuned models to improve the execution module's ability to follow planning instructions;
\item We conduct Routine-based knowledge distillation to generate a scenario-specific multi-tool dataset, which improves the accuracy of the execution model.
\end{enumerate} 
These results demonstrate the effectiveness of Routine in enhancing the stability and accuracy of agent systems in solving complex problems within specific scenarios. 

\section{Related Work}

\subsection{Agent Planning and Execution Framework}
LLMs and agents have introduced potential methods for enterprises to improve operational efficiency, particularly by enabling intelligent automation and decision support~\cite{muthusamy-etal-2023-towards}. The initial wave of LLM-based agent practice from 2023 to 2024 was marked by the emergence of several exploratory systems such as AutoGPT, a task list–driven autonomous agent~\cite{yang2023autogptonlinedecisionmaking}, BabyAGI, a task manager capable of dynamic goal generation and prioritization~\cite{talebirad2023multiagentcollaborationharnessingpower}, and AutoGen, a multi-agent framework that leverages GraphFlow to model execution as a DAG of tool calls, ensuring traceable and reliable process flows~\cite{wu2023autogenenablingnextgenllm}. Collectively, these systems underscored the feasibility of using LLMs for autonomous reasoning, task decomposition, and multi-step tool orchestration. At the same time, these projects highlighted limitations in early architectures, including inefficiencies in context management~\cite{chang2025sagallmcontextmanagementvalidation}, fragmented execution flows~\cite{chang2025sagallmcontextmanagementvalidation}, and variability in task completion success rates~\cite{qi2025agentifbenchmarkinginstructionfollowing} -- particularly stemming from step explosion and substantial variability in running cost caused by frequent execution loops~\cite{kim2025costdynamicreasoningdemystifying, lu2025runawayashamedhelpfulearlyexit}.

In response to these limitations, a second wave of LLM-based agent development emerged in 2025, shifting the research focus to improve agent's stability, reliability, and enterprise applicability.
Representative systems adopted production-oriented agent architectures featuring modular and collaborative designs~\cite{pokhariya2024manusmarkerlessgraspcapture, zhang2025agentorchestrahierarchicalmultiagentframework}. These systems followed a plan-then-act paradigm, wherein high-level task planning precedes the dynamic selection and execution of tools for individual subtasks~\cite{yao2022react, erdogan2025planandactimprovingplanningagents}. This paradigm preserves execution flexibility while discipline planning, marking a significant step toward practical and controllable agent systems for real-world deployment.
For example, Manus employs a three-stage pipeline, Planner, Executor and Verifier, to ensure accuracy and traceability~\cite{pokhariya2024manusmarkerlessgraspcapture}.

Despite these advancements, many existing systems still represent plans in unstructured or semi-structured natural language, which hampers verification, debugging, and reuse of process~\cite{zheng2024naturalplanbenchmarkingllms, gestrin2024nl2planrobustllmdrivenplanning}. In practice, this introduces ambiguity during execution, hinders static validation, and limits the agent's ability to track intermediate states or recover from failure. To address these challenges, we propose Routine —- a structured planning script that serves as an intermediate representation between LLM-generated plans and execution engines, enhancing the execution model's instruction-following capability.

\subsection{Instruction-Following Capabilities of LLMs}
The increase in model size emerges the in-context learning abilities of LLMs~\cite{Diao2025}. By providing task-specific rules and output constraints in the system prompt, an LLM can interpret instructions and adapt to concrete tasks. As prompt engineering reveals substantial practical value, researchers increasingly focus on instruction-following, giving rise to specialized training schemes and evaluation protocols~\cite{wei2023chainofthoughtpromptingelicitsreasoning, sanh2022multitaskpromptedtrainingenables, wang2023selfinstructaligninglanguagemodels}.

Google DeepMind introduced \textit{IFEval}~\cite{Zhou2023}, which provides 25 verifiable instructions and 500 prompts, providing a standardized baseline. However, its coverage of task chains, state dependence, and composite constraints remains limited. To transcend this flat evaluation surface, Tsinghua University proposed \textit{ComplexBench}~\cite{Wen2024}. By combining chain, parallel, and nested structures across 19 constraint dimensions, \textit{ComplexBench} shows that GPT-4's accuracy drops markedly on chain structures and deeply nested choices, highlighting weaknesses in structural awareness and distributional generalization.

To strengthen alignment under complex constraints, Tongyi Lab proposed \textit{IOPO} (Input–Output Preference Optimization) and released the \textit{TRACE} benchmark~\cite{Zhang2024}. IOPO jointly models preference pairs from instruction and response, yielding significant gains on difficult instruction-following tasks. ByteDance built \textit{GuideBench}~\cite{Diao2025}, introducing \textit{guideline rules} that emulate dynamically evolving domain regulations. Experiments show that mainstream LLMs still struggle with fine-grained, domain-specific rules. As an alternative, context engineering offers a lightweight way to improve model compliance by structuring prompts and adding task-relevant cues. However, manual designs remain impractical and struggle with deeply structured or branching tasks~\footnote{https://github.com/davidkimai/Context-Engineering}.

Routine was first introduced in the OpenAI Cookbook~\cite{openai_cookbook_using_reasoning_2024}. According to the Cookbook, Routine enables the decomposition of instructions into smaller, manageable tasks, thereby reducing the risk of hallucinations in LLMs and facilitating more effective customer service solutions. in our research, we added more components to Routine and further optimized its structure, and we also focused more on the model's \textit{procedural instruction-following ability}. By providing the model with a structured Routine format that explicitly encodes the entire workflow, we improve its stability in following the prescribed plan. Constructing a training set that pairs inputs with this Routine representation further enhances the model's ability to generate outputs that conform to the required structure.

\subsection{Tool Calling Data Synthesis and Post-Training}
Training data quality and structure are crucial to the performance of LLM. However, the collection of high-quality, human-annotated scenario-specific datasets entails substantial labor costs. A common strategy to tackle this issue is to prompt LLMs to synthesize training data, which is subsequently used to enhance model performance through targeted post-training~\cite{guo2025synthetic}. In tool calling, data synthesis pipelines are essential for consistency and effectiveness. To this end, several studies have proposed dataset generation frameworks~\cite{qin2023toolbench, liu2024toolace, tang2023toolalpaca, liu2024apigen}. Beyond these efforts, \textit{Kimi-k2} introduced a scalable automated pipeline to simulate real-world multi-turn tool-use scenarios and generate large-scale, diverse, and high-quality training datasets~\cite{kimi-k2}. 
These frameworks primarily focus on generating tool-calling datasets to improve models' ability to read inputs and produce standardized tool calling commands, ensuring consistent output of tool calling instructions aligned with commonly accepted structural conventions in the ecosystem.

Despite these advances, common tool-use models still under-perform in scenario-specific scenarios. While they maintain consistent output formatting, they often struggle to select the correct functions or parameters due to a lack of scenario-specific knowledge~\cite{guo2025synthetic}. Prompt engineering and constructing scenario-based training datasets using knowledge distillation can be used to address the issue~\cite{hinton2015distilling}.
Distilling expert knowledge about tool selection and parameter configuration into smaller models enables lightweight expert agents to solve scenario-specific tasks.

Building on prior work in enterprise-level tool calling, we previously developed a training pipeline to improve single-step function execution by generating task-specific data based on a curated tool list~\cite{zeng2024adaptablepreciseenterprisescenariollm}. The present study extends that work by focusing on multi-step tool calling capabilities.\\

\section{Agent System Framework}

\begin{figure}
    \centering
    \includegraphics[width=1\linewidth]{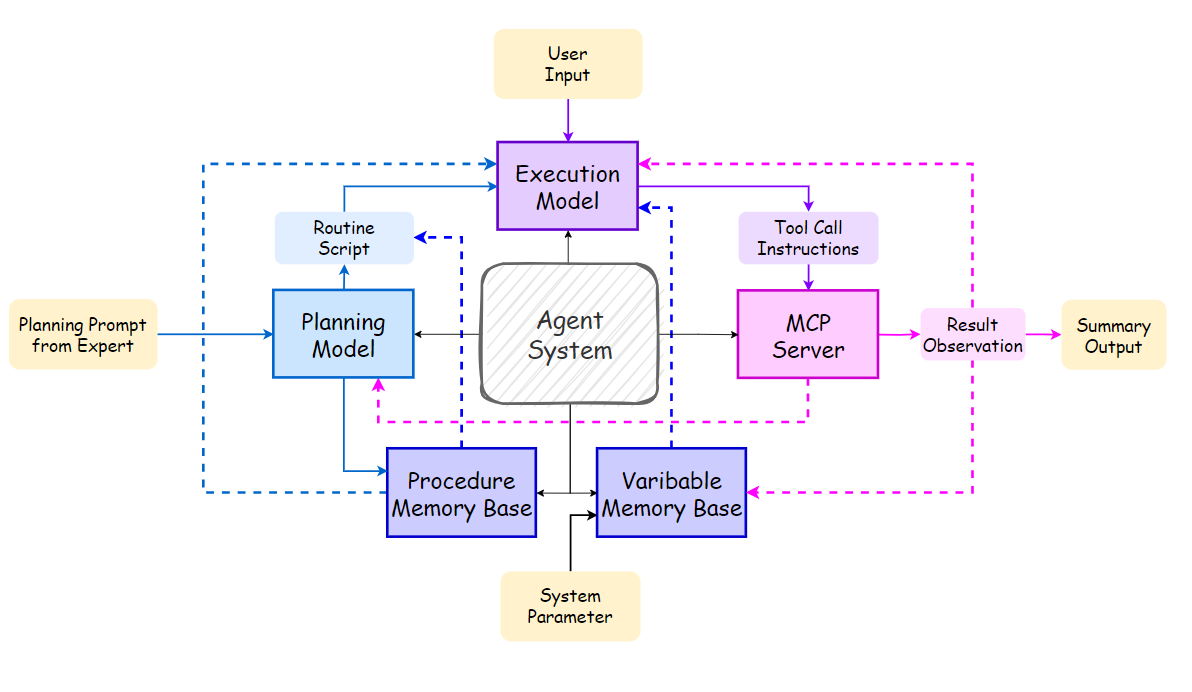}
    \caption{Framework of our Routine-based agent system. To begin, the system uses expert-annotated planning prompts to generate Routines. During runtime, it processes user input and system parameters, completing tasks through the interaction between the modules, and ultimately provides a final summary based on the observation from a dedicated summarization tool.} 
    \label{fig:agentStructure}
\end{figure}

An agent system typically comprises four core modules that collaborate on sole user tasks: \textbf{Planning, Execution, Tools, and Memory}~\cite{Wang_2024}.
When a task is received, the \textbf{Planning Module} generates a step-by-step plan. The \textbf{Execution Module} then follows this plan, generating the appropriate tool call instruction for each step. The \textbf{Tool Module} receives these instructions and returns the execution results~\cite{Qu_2025}. The \textbf{Memory Module} assists the entire agent, improving execution coherence and enhancing generation. Based on our Routine and other optimization mechanisms, we have redesigned these four primary modules to create an agent system centered around Routines and multi-step tool calling mechanism. The overall architecture of this system is shown in Figure \ref{fig:agentStructure}. In this section, we will introduce how we design these functional modules.

\subsection{Planning Module and Routine}
 In our system, the main purpose of Routine is to enhance the stability of common models in handling scenario-specific tasks, thereby improving the model's instruction-following capabilities to increase execution accuracy.
 \begin{table}[htbp]
\centering
\caption {The components of a Routine. \small Note: Components marked with * are optional elements. }
\vspace{+0.3em}
\label{tab:sub-task-steps}
\newcommand{\rowheight}{1.7em}

\begin{tabular}{|m{3cm}|m{10cm}|}
\hline
\parbox[c][\rowheight][c]{\linewidth}{\textbf{Component}} & 
\parbox[c][\rowheight][c]{\linewidth}{\textbf{Description}} \\ \hline
\parbox[c][\rowheight][c]{\linewidth}{Step Number} & 
\parbox[c][\rowheight][c]{\linewidth}{The sequential number of the step within the main process.} \\ \hline
\parbox[c][\rowheight][c]{\linewidth}{Step Name} & 
\parbox[c][\rowheight][c]{\linewidth}{A concise summary of the step's purpose or function.} \\ \hline
\parbox[c][\rowheight][c]{\linewidth}{Step Description} & 
\parbox[c][\rowheight][c]{\linewidth}{Detailed instructions, execution conditions, and objectives of the step.} \\ \hline
\parbox[c][\rowheight][c]{\linewidth}{Input Description*} & 
\parbox[c][\rowheight][c]{\linewidth}{A description of the parameters required for executing this step.} \\ \hline
\parbox[c][\rowheight][c]{\linewidth}{Output Description*} & 
\parbox[c][\rowheight][c]{\linewidth}{The output parameters generated after successful execution.} \\ \hline
\parbox[c][\rowheight][c]{\linewidth}{Step Tool*} & 
\parbox[c][\rowheight][c]{\linewidth}{The corresponding function tool used in this step; only one tool is called.} \\ \hline
\end{tabular}

\end{table}

\subsubsection{Routine Components and Format}
A Routine is composed of multiple specific execution steps of sub-tasks, which are independent yet often interrelated. Therefore, a Routine execution sequence must contain sufficient information for the agent to reliably follow the planned steps. The components of a complete Routine and their descriptions for a sub-task sample are described in Table ~\ref{tab:sub-task-steps}.

Routines for similar scenarios may contain overlapping steps, differing only in certain process segments, similar to different branches of the same workflow. In such cases, similar scenarios can be merged by creating branching steps and executing conditions to differentiate between the workflows in one single Routine. 

\noindent Here is an example of a Routine with branches:

\begin{quoteBox}
\begin{quote}
\footnotesize
\textbf{Step X.} \textless Step Name\textgreater: This step performs a branch condition check:

\begin{itemize}
  \item \textbf{Branch X-1 Step 1.} \textless Step Name\textgreater: If \textless Condition\textgreater, perform \textless Step Description\textgreater, using the \textless Tool Name\textgreater\ tool;
  \item \textbf{Branch X-1 Step 2.} \textless Step Name\textgreater: \textless Step Description\textgreater, using the \textless Tool Name\textgreater\ tool;
  \item \textbf{Branch X-2 Step 1.} \textless Step Name\textgreater: If \textless Condition\textgreater, perform \textless Step Description\textgreater, using the \textless Tool Name\textgreater\ tool;
\end{itemize}

\textbf{Step Y.} \textless Step Name\textgreater: \textless Step Description\textgreater, using the \textless Tool Name\textgreater\ tool;\\  
\textbf{Step Z.} \textless Step Name\textgreater: \textless Step Description\textgreater, using the \textless Tool Name\textgreater\ tool, and terminate the workflow;
\end{quote}
\end{quoteBox}

For instance, a branch can be denoted as "Branch X-n Step i", representing the \textit{i}-th step within the \textit{n}-th branch of main step \textit{X}. Based on this notation, the execution model can determine its current branch and position within the workflow. When a workflow terminates after a certain step, this must be described in the Routine accordingly. This structured format ensures that each planning step contains clear and complete information, facilitating progress tracking and task execution for the agent. When a workflow needs to be modified, developers can quickly insert, delete, or edit steps, allowing for agile adjustments to the task plan. An example of a Routine can be found in the Appendix~\ref{sec:apx_b}.

\subsubsection{AI-Powered Routine Generation and Optimization}
While general-purpose models can effectively deconstruct problems into natural language Routines in open-ended scenarios, they struggle in enterprise contexts due to a lack of domain-specific knowledge. This makes it difficult for them to directly and reliably generate contextualized Routines without assistance from domain experts or enterprise documentation.

To improve efficiency, an expert can provide a simple draft prompt outlining the plan of a specific scenario. This draft is then optimized by a model equipped with a specialized prompt template. The optimization process involves decomposing the plan into detailed sub-steps, mapping these steps to available tools, and finally outputting a structured and comprehensive natural language Routine that is easy for the execution module to follow. The overall processing flow is shown in Figure \ref{fig:AIRoutineOpt}. The prompt template for this AI-driven optimization can be found in the Appendix~\ref{sec:apx_a}.

\subsection{Execution Module and Small-Scale LLMs}
The agent's \textbf{Execution Module} is responsible for receiving the plan provided by the Planning Module and following its prescribed path to output tool calling instructions. In most agent systems, both planning and execution are handled by the same model. Due to the diversity and complexity of tasks, planning often requires high performance and consumes significant computational power and inference time, requiring the use of large-scale models \cite{neelakrishnan2024}.

However, in enterprise-level agent systems, the execution process brings a large amount of context, including numerous decision nodes and available tools. Using a high-performance model for each execution step consumes substantial resources and time, making it difficult to deploy in a real enterprise environment \cite{guguloth2025}.

For the Execution Module, the core required capabilities are multi-step tool calling and instruction following, rather than complex logical reasoning or abstraction. Therefore, the Execution Module can be driven by a smaller, specialized instruction-following model. This model is only responsible for following the plan from the planning model and outputting the corresponding tool calling instructions. The execution model does not generate natural language responses; instead, a final summary is generated by a dedicated summarization tool in the last step \cite{zeebaree2025}. This separation ensures that the prompt templates for summarization and tool calling do not interfere with each other.

By providing a predefined plan, an Execution Module powered by a small specialized model can save significant resources, thus improving its viability for real-world enterprise agent applications \cite{gopalaswamy2025}.

\begin{figure}
    \centering
    \includegraphics[width=1\linewidth]{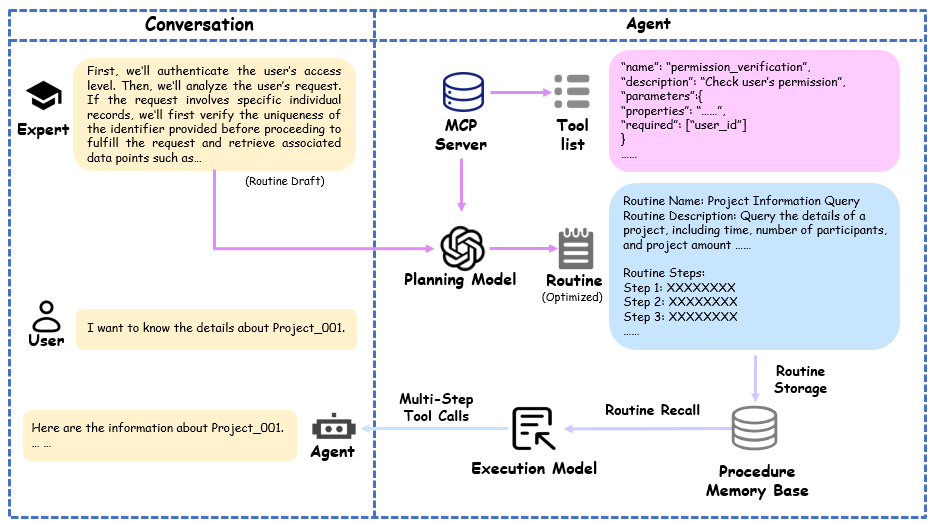}
    \caption{AI Optimization and Management of Routines}
    \label{fig:AIRoutineOpt}
\end{figure}

~\subsection{Tool Module and MCP Server}
The Tool Module is responsible for receiving tool calling instructions from the Execution Module, using these tools to perform tasks, and returning the execution results, thereby providing the agent system with external information. In our system, we use MCP servers as the Tool Module. The MCP server not only serves as the resource for the Execution Module but also defines and manages the set of tools available to the agent ~\footnote{https://www.anthropic.com/news/model-context-protocol}.

MCP defines tools in a structured manner with its protocol. Each tool is uniformly described by this protocol, which specifies its name, parameter types, and calling constraints. The execution module simply retrieves the target function from the MCP tool set by step-by-step instructions from the Planning Module and fills in the required parameters. This protocol-based structure no longer needs to manage the implementation details of the tools. Consequently, the Execution Module only needs to determine "which tool to call" and "what parameters to pass" based on the plan provided by the Planning Module~\cite{gao2025mcp}.

Furthermore, the standardized design of the Tool Module provides the system with extremely high extensibility, making it easy for developers to add new functions or connect with new systems. This allows for a diverse ecosystem of tools, including various functions, models, other agents, or even user responses to assist the agent.

In enterprise scenarios with a wide variety of complex and interactive tools, the MCP Tool Module effectively decouples the execution logic from the tool layer, creating a clear division of labor: the Tool Module provides stable and reliable function standards and interfaces, while an Execution Module driven by a small-scale model calls these tools step-by-step~\cite{gopalaswamy2025}.

\subsection{Memory Module: Procedures and Variables}

Agent systems typically process a large volume of context, including system prompts, lists of available tools, and conversation history~\cite{ji2025surveyprogressllmalignment}. This imposes a significant load on the model: increased context not only leads to higher inference costs but also decreases the model's attentional accuracy. This requires a specialized memory storage and retrieval mechanism that provides the agent with only the most relevant information for the current task~\cite{kate2025longfuncevalmeasuringeffectivenesslong}.

In our agent system, we have implemented two forms of memory: long-term \textbf{Procedure Memory} and short-term \textbf{Variable Memory}.

\subsubsection{Procedure Memory}

Through the AI-driven Routine generation and optimization process, the Planning Module creates a collection of Routines needed for a given scenario in collaboration with domain experts. Each group of Routines in this collection is designed to handle a specific sub-task. Since a single scenario may contain multiple sub-scenarios and their corresponding Routines, placing this entire collection into the execution model's system prompt would significantly increase inference costs and could decrease accuracy by introducing irrelevant information \cite{neelakrishnan2024}.

Therefore, we established a Procedure Memory base for our agent system. Before deployment, experts populate this memory base with the necessary set of Routines. When the system receives a relevant query, it retrieves the appropriate Routine(s) from memory based on a similarity calculation between the Routine's description and the user's task, thereby assisting the execution model \cite{guguloth2025}.

\subsubsection{Variable Memory}

In multi-step tool calling processes, the execution history gradually accumulates. The input and output parameters often occupy a large portion of the context window, leading to issues such as excessively long parameter values, excessive number of parameters, and redundant punctuation. These problems not only increase the pressure on the model's context window and heighten the probability of model hallucinations, but also cause smaller models to make syntax errors involving brackets, quotes, and escape characters when passing parameters~\cite{kate2025longfuncevalmeasuringeffectivenesslong}.

\begin{wrapfigure}[24]{r}{0.5\linewidth} 
    \centering
    \includegraphics[width=0.95\linewidth]{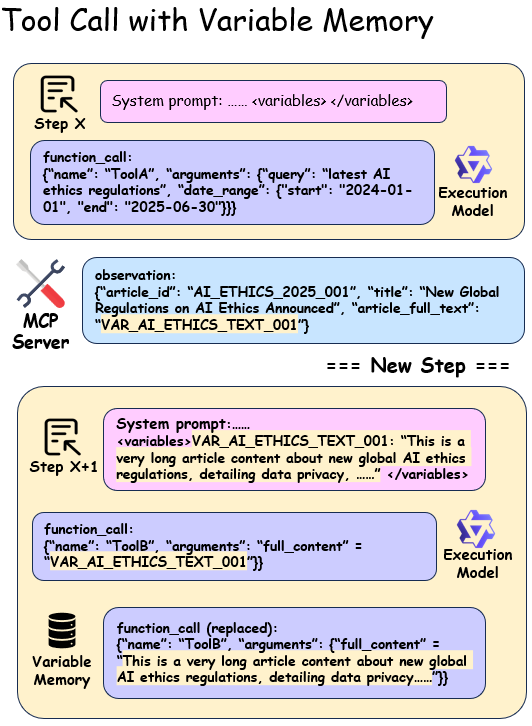}
    \vspace{-0.5em}
    \caption{A schematic diagram of agent's variable memory mechanism.} 
    \label{fig:varMemory}
\end{wrapfigure}

To effectively address this issue, we introduce our \textbf{Variable Memory} mechanism, as shown in Figure \ref{fig:varMemory}. Its core function is to optimize parameter passing between multi-step tool calls. When a tool call returns an excessively long parameter, the system automatically stores it in the variable memory base. The model then only needs to provide the corresponding key when filling in the tool parameters, rather than the full value. Upon receiving the tool call request, the Memory Module automatically retrieves these keys back to their actual parameter values before passing them to the tool~\cite{kate2025longfuncevalmeasuringeffectivenesslong}. The Variable Memory mechanism significantly reduces context pressure, and it also reduces token consumption and improves syntactic accuracy.

It is important to note that this variable memory is non-persistent; all related variable memories are used only for the execution of the current task. This design ensures that memory remains lightweight and responsive, avoids unnecessary storage overhead and data accumulation, and guarantees the independence of each task execution ~\cite{lin2025evaluatinglargelanguagemodels}.\\

\section{Agent Execution Model Training}

Common models can effectively execute Routines based on their generalized prompt-following capabilities. However, even when the tools and parameters for each step are specified, lightweight models remain susceptible to hallucinations, often leading to incorrect tool calls and parameter updates \cite{neelakrishnan2024}. To address this, a common Routine-following dataset can be constructed to enhance the execution model's ability to follow instructions. In addition, Routines can serve as the procedural knowledge for data distillation: By using high-capacity models and Routine, we can generate scenario-specific multi-step tool call training data. This distilled data can then be used to train scenario-specific execution models, thereby reducing their reliance on explicit planning \cite{guguloth2025}. 

\begin{figure}
    \centering
    \includegraphics[width=1\linewidth]{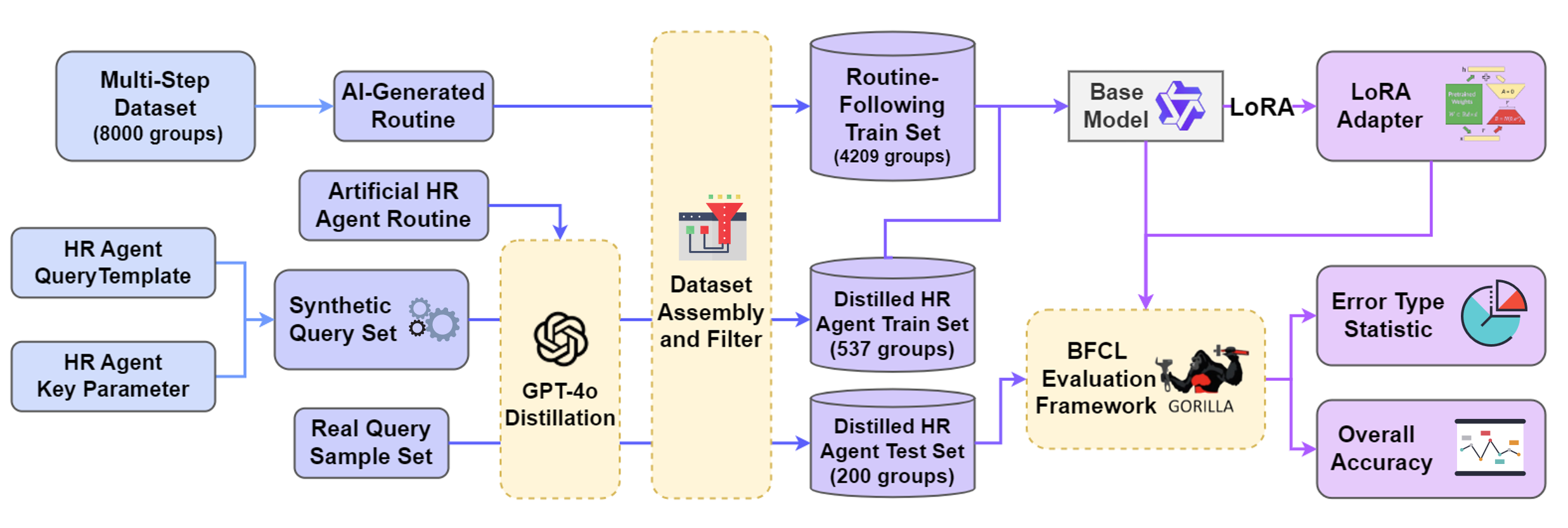}
    \caption{The agent executes the model training process. The whole process includes the common Routine following capability training and the scenario-specific tool calling capability training, and can be divided into three modules: data synthesis, model fine-tuning and model evaluation.}
    \vspace{-0.5em}
    \label{fig:agentModelTrainPipeline}
\end{figure}

In this paper, we explore the impact of these two training strategies on improving the capabilities of execution models, with the aim of identifying the best ways for leveraging Routine-based planning in real-world scenarios \cite{zeebaree2025}. The complete process of model training from data synthesis to evaluation is briefly shown in Figure \ref{fig:agentModelTrainPipeline}.

\subsection{Experimental Scenario Setting and User Queries}
In this experiment, we selected a real-world enterprise scenario as the test setting: an HR agent application deployed within a large corporation (with more than 8,000 employees). In this scenario, HR users interact with the agent to inquire about information related to employees and departments within the corporation.~\cite{zeng2024adaptablepreciseenterprisescenariollm}

The HR agent scenario comprises seven sub-scenarios, each guided by a distinct Routine used to instruct the execution model in completing user queries, thereby requiring seven corresponding Routines. In this experiment, we manually annotated all seven Routines. Each Routine consists of 4–7 tool calling steps, with each tool requiring 1–3 input parameters. Due to parameter dependencies across tools, the execution order of steps within a Routine must follow a specific sequence. The agent has access to an MCP server that contains 25 functionally distinct tools with different functionalities, including data query, permission verification, and model generation. Some tools require system parameters such as current user ID, which are provided in the system prompt prior to task execution. Notably, 3 out of the 7 Routines in the evaluation scenario involve branching logic. During execution, the agent must evaluate the specified conditions and follow the appropriate branch to complete the task. If these branching Routines are further decomposed into non-branching ones, the HR agent system scenario can be represented as 10 distinct unbranched Routines.

In the final summarization step, Routine guides the execution model to call a summarization tool, rather than directly generating natural language output by the execution model itself. This tool is paired with a scenario-specific summarization template to ensure that the output remains unaffected by other components of the system prompt. Once the execution model calls the summarization tool, the multi-step tool-calling process is complete, and the tool's output is returned directly to the user.

\subsection{Data Synthesis}
\subsubsection{System Prompt Template for Execution Model}
The system prompt template for synthetic data must incorporate all necessary information required by the execution model to resolve a task. It includes Routine-specific content such as role definitions, task, and behavioral guidelines, along with system parameters, a Routine tailored to the problem, a variable memory dictionary, and a tool list -- all organized in precise order as the model's operational context. A representative example of such a system prompt is provided in the Appendix~\ref{sec:apx_a}.

\subsubsection{Common Data Synthesis}
We synthesized generalized Routines from the training data using the BUTTON open-source dataset (BUTTONInstruct)\footnote{https://github.com/PKU-Baichuan-MLSystemLab/BUTTON}, aiming to enhance the model's robustness in Routine-following across diverse scenarios, rather than focusing solely on the procedural steps of a single case. The BUTTON dataset comprises 8,000 single-turn, multi-step tool call instances spanning a broad array of common task types. Each instance alternates between tool\_call and observation to form a clearly structured execution trajectory. Leveraging these records, we utilized GPT-4o  with a specialized prompt template to generate precise and structured Routines: each Routine enumerates the step index and name, functional objectives, and the corresponding tool name. We then constructed new system prompts based on a standardized execution prompt template to improve adaptability to various workflow mechanisms.

Subsequently, we performed targeted data filtering to optimize training efficiency and model performance. Specific optimizations are illustrated in Figure \ref{fig:datafilt} and include:

\begin{figure}[H]
    \centering
    \includegraphics[width=0.85\linewidth]{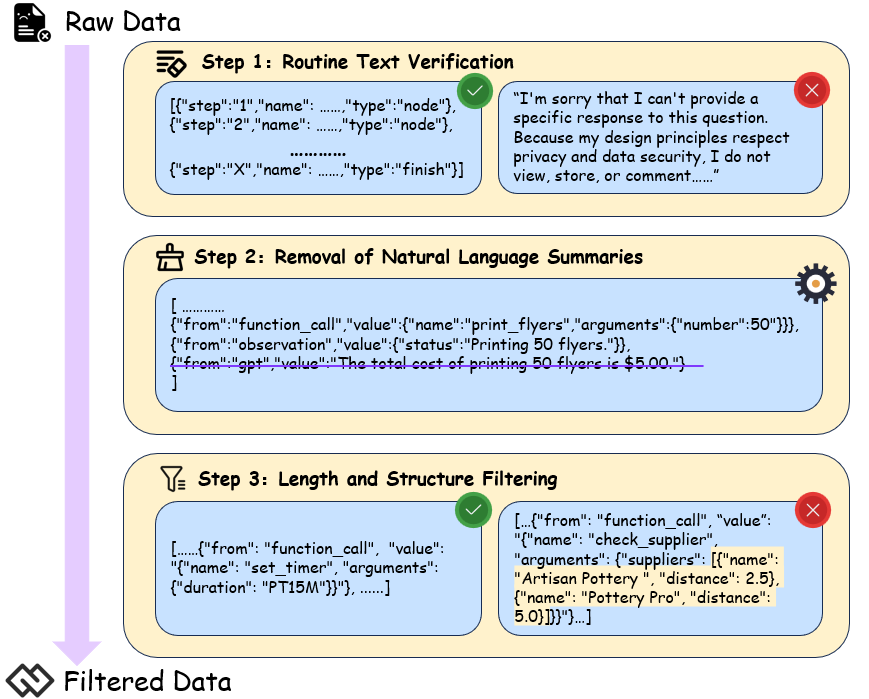}
    \caption{The common Routine-following dataset filtering pipeline.}
    \vspace{-1.8em}
    \label{fig:datafilt}
\end{figure}
\vspace{1.0em}

\begin{enumerate}
\item \textbf{Routine Text Verification:} Basic verification of generated Routine texts, discarding entries with empty responses or abnormal outputs.

\item \textbf{Removal of Natural Language Summaries:} As final agent-level summaries are delegated to a specialized summarization tool, we removed all natural language summaries steps, retaining only user queries, tool calls, and corresponding observations to stabilize tool calling outputs.

\item \textbf{Length and Structure Filtering:} We excluded instances with more than eight tool call steps to prevent excessive computational load and removed any instances containing nested lists, dictionaries, or other complex data structures to maintain format simplicity and symbolic accuracy.
\end{enumerate} 

Based on these filtering procedures, we retained 4,209 high-quality, lightweight training data, which aim to strengthen the model's capability for following Routine in different specific scenarios.

\subsubsection{Scenario-Specific Routine Knowledge Distillation}
Besides serving as system prompts to provide scenario knowledge to execution models, Routines can also be used as materials for knowledge distillation~\cite{kujanpää2024knowledgeinjectionpromptdistillation}. Based on user queries within specific scenarios, Routines can be incorporated into the system prompt of a teacher model to generate stable, multi-step tool calling records. These outputs can then be used to train a lightweight student model, enabling the agent to complete multi-step tool calls within a scenario without relying on explicit Routine. 

To create a set of user queries in the HR agent scenarios, we designed 5–6 user query templates for each of 10 distinct, non-branching sub-scenarios. By filling out these templates with various department and employee information, each sub-scenario yielded approximately 50–60 unique user queries. After data cleaning, we obtained a total of 537 user queries. To further enhance data diversity and improve the generalization ability of the model, we performed data augmentation by using LLM to generate semantically equivalent variations with different phrasing, which improves the model's robustness.

Then, we used GPT-4o equipped with Routines to distill a set of 537 single-turn, multi-step user queries for the HR agent system scenario after data cleaning. Each sample includes 4–7 tool calling steps, totaling 3108 labeled tool calling instructions, which were used to train the execution model in a specific-scenario. In addition, we distilled a set of 200 single-turn, multi-step scenario-based user queries, totaling 1,148 tool calls. This dataset was used to evaluate model performance across different variants.

\subsection{Model Training}

\subsubsection{Base Model Selection and Setting}
In the HR agent scenario, the execution model needs to have a certain level of instruction-following and preliminary tool-calling capability, as well as a strong Chinese understanding. In addition, the execution model requires a small number of parameters, maintaining the efficiency of the agent system in terms of computing resources and time consumption. Therefore, we select qwen2.5 and qwen3 series of small-scale Instruct models for training. When training the qwen3 series models, we did not enable model reasoning training, because the training goal is mainly to improve the model's ability to follow Routine, and there is no need to utilize thinking to infer the tool calls instruction, thereby saving the tokens of model reasoning.

We added <routines>, </routines>, <variables>, </variables> to the special tokens to append Routine fields and store variable memories. Registering them in the vocabulary can make them effectively recognized by the model and improve the stability of model training. 

\subsubsection{LoRA Fine-tuning Setting}
To avoid overfitting and maximize ROI under enterprise-level computational constraints, we adopted LoRA, a lightweight fine-tuning strategy, aiming for cost-efficient adaptation on small-scale datasets~\cite{hu2021lora}.

Our experiments utilized the LLaMA-Factory framework with DeepSpeed ZeRO-3 and Flash Attention-2, which can maximize computational efficiency and significantly reduce GPU memory usage during multi-GPU training~\cite{rajbhandari2020zeromemoryoptimizationstraining}~\cite{dao2023flashattention2fasterattentionbetter}. To enhance the model's ability in comprehending and reasoning over structured data in business scenarios, we adjusted the maximum sequence length to exceed the actual input length in all cases, thereby ensuring the preservation of critical structural information like multi-step tool call chains during preprocessing.

We created two fine-tuning datasets: a scenario-specific HR agent dataset containing 537 instances and a general dataset with 4,209 filtered instances. Model training was performed on four NVIDIA A10 GPUs (24GB VRAM each), with a LoRA rank of 8, batch size of 1 per GPU, and gradient accumulation steps set to 4, resulting in an effective batch size of 16 per update. The learning rate was set to 1.0e-4 with a warm-up ratio of 0.1. Based on validation performance, the final model was selected from the checkpoint at epoch 3, striking a balance between scenario-specific adaptation and generalization capability.\\

\section{Evaluation and Analysis}
\subsection{Evaluation Methodology}
\subsubsection{Evaluation Framework and Procedure}
To comprehensively and automatically evaluate the execution model's ability to handle multi-step tool calls, we choose the open-source Berkeley Function-Calling Leaderboard (BFCL) as our core evaluation framework, primarily based on its Function-Calling (FC) mode and Abstract Syntax Tree (AST) evaluation method. It offers the dual advantages of evaluation efficiency and precise error attribution: its speed is not affected by tool response latency, while it also provides a detailed analysis of various error sources in the model's tool calling instructions~\cite{patil2025bfcl}.

The AST evaluation process follows a hierarchical order, and we further categorize the results into three main evaluation metrics to calculate individual accuracy rates and overall accuracy:
\begin{itemize}
\item \textbf{Structural Error:} The model output is first checked for valid JSON formatting. Errors include missing brackets, punctuation errors, and other issues that could cause AST parsing to fail. 

\item \textbf{Tool Selection Error:} If the structure is valid, the tool choice is evaluated. Errors include outputting natural language instead of using a tool, calling incorrect numbers of tools, or wrong tools (e.g., confusing similar tools or nonexistent ones).

\item \textbf{Parameter Error:} Includes three subcategories: incorrect parameter values, parameter hallucination (filling in parameters that do not exist in the tool definition, and missing key parameters.) Note: This does not cover the detection of content from a free text parameter.

\item \textbf{Overall Accuracy:} A case is counted as correct only if all steps (structure, tool, and parameters) are entirely accurate. Overall accuracy is the most stringent measure of the model's end-to-end problem-solving ability.
\end{itemize} 

\begin{figure}
    \centering
    \includegraphics[width=1\linewidth]{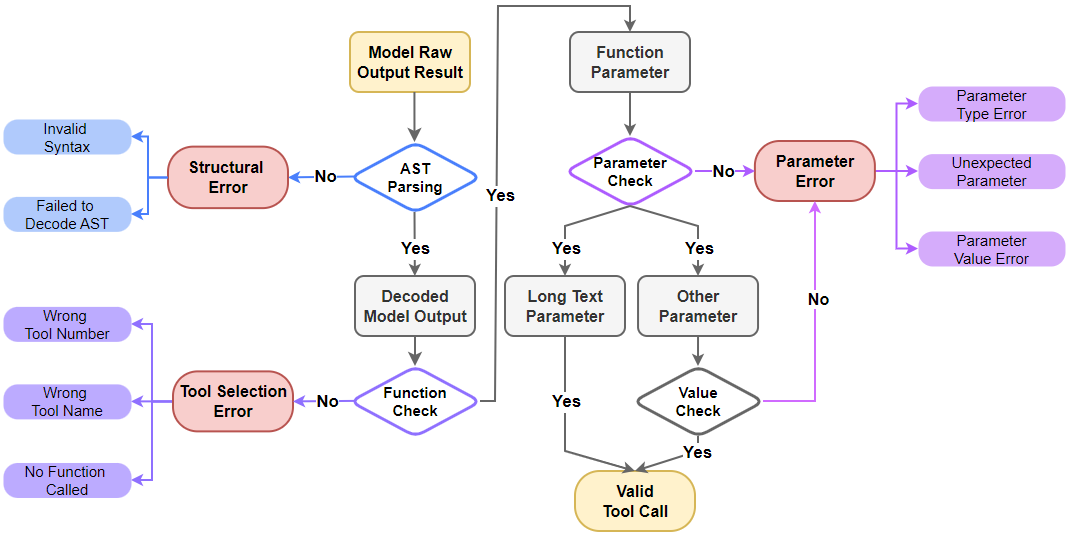}
    \caption{AST evaluation workflow for our enterprise scenario. The process is mainly based on the BFCL framework with categorizing errors into three primary types: structural, tool selection, and parameter errors. Note: The content of free text parameters is not assessed via exact matching during parameter evaluation.}
    \vspace{-0.5em}
    \label{fig:bfcl}
\end{figure}

The evaluation procedure is hierarchical as shown in Figure ~\ref{fig:bfcl}. The model output is prioritized for structural accuracy since a structural failure prevents the parsing of tools and parameters. The selection of tools is then assessed. Finally, the parameter correctness is judged only if the previous steps are successful. This hierarchy is also reflected in our statistical analysis: the structural accuracy is calculated on all samples, the tool selection accuracy is calculated on the structurally correct subset, and the parameter accuracy is calculated on the subset where both structure and function calls were correct. This ensures an objective measure of the actual accuracy for each category.
We especially implement approximate matching for free text parameters that are frequently passed in this enterprise scenario. Rather than checking the exact value of parameters, it focuses on verifying parameter existence, illusion, and types. Slight phrasing differences in long natural language parameters (e.g. \textit{user need-query} v.s. \textit{user need: query}) do not significantly alter the semantic meaning but would fail the strict-matching test, leading to an underestimation of the model's core abilities. The approximate matching allows the evaluation system to focus more on the model's performance in instruction structure, tool selection, and key-parameter passing, providing a more objective reflection of its comprehensive tool calling capabilities.

\subsubsection{Test Data Synthesis}

We constructed a specialized evaluation dataset to accurately assess the model's performance. Using the model distillation, we generated a 200 single-turn, multi-step evaluation dataset from the HR agent scenarios. The data were then decomposed into 1,148 individual test samples for our automated evaluation framework. Based on optimizations to the execution module, we focus on two key principles to ensure fairness and validity during the decomposition process:

\begin{enumerate}
\item \textbf{Decomposition of Test Samples:} Each full execution trace from the distillation process was split into multiple, independent test samples. Every sample includes the history of execution steps preceding the current one, as well as the specific system prompt for the current step.

\item \textbf{Records of System Prompt:} Due to the mechanism for storing temporary memory variables, the system prompt received by the agent can be dynamically updated at each execution step. We store the state of the system prompt at every step to fully preserve the agent's history context. This ensures that the model is evaluated with the exact context it would have in a real continuous interaction, which is essential for accurate analysis of the model's tool call ability and related errors~\cite{ye2025taskmemoryengine}
\end{enumerate} 

Furthermore, we randomized the sequence of tools in the list provided for each test sample to ensure that the model demonstrates a true understanding of the tools rather than relying on their order. This prevents the model from depending on positional biases and helps to guarantee the generalization ability of our findings~\cite{guan2025evaluatingllmbasedagentsmultiturn}.

\subsubsection{Test Configuration}
To comprehensively examine the model's performance under different Routine configurations, we generate three distinct evaluation scenarios from the original 1,148 test samples by modifying the Routine component:

\begin{enumerate}
\item \textbf{No-Routine Scenario (Baseline):} The execution model only receives the user's query and must autonomously understand the intent, devise a plan, and complete the tool calls. This serves as our baseline for evaluating the model's inherent capabilities and for comparison against optimization effects.

\item \textbf{Routine-Guided Scenario w/o Branches:} The model is provided with a structured, linear Routine in natural language that outlines a clear path to the solution without any complex conditional branches. This tests the model's instruction-following capability when given explicit linear instructions.

\item \textbf{Routine-Guided Scenario w/ Branches:} This scenario also provides the model with the same Routine, but it includes conditional branches that require the model to make judgments and selections based on the results of intermediate steps. It is designed to evaluate the model's execution stability and logical reasoning when handling more complex non-linear workflows.
\end{enumerate}  

In these tests, the Routine consists of the step number, name, description, and tool, but does not include input/output parameter descriptions. The impact of different Routine components is further explored in the following ablation studies.

\subsection{Results and Discussion}
For this experiment, we selected a variety of leading industry foundation models for testing, including proprietary models: GPT-3.5-Turbo, GPT-4-Turbo, GPT-4o, and Claude-3.7-Sonnet, as well as the original versions of open-source Qwen series models: Qwen-2.5-7b-Instruct, Qwen-2.5-14b-Instruct, Qwen-3-8b, and Qwen-3-14b. The experiment also evaluated the effects of training for common Routine following and the impact of training on multi-step tool calling data that was distilled using Routine. The model training strategies are defined as follows:

\begin{itemize}
\item \textbf{Common Routine following fine-tuning:} Based on 4,209 samples synthesized common training dataset, aim to improve the model's generalized ability to follow the structured Routine.

\item \textbf{Scenario-specific tool calling fine-tuning:} Based on 537 single-turn, multi-step tool calling data samples distilled from the target scenario, aim to directly improve the model's ability to use the scenario's tools for multi-step tool calls.
\end{itemize}
\begin{table}[htbp]
\centering
\renewcommand{\arraystretch}{1.4}
\caption{Overall accuracy of models under different Routine configurations in HR agent system scenario. The fine-tuned models are categorized by their training dataset: \textit{Common} refers to models trained on the common Routing-following dataset, while \textit{Scenario} refers to models trained on the distilled multi-step tool calling dataset.}
\label{tab:models-performance-on-routine}
\resizebox{\textwidth}{!}{%
\begin{tabular}{l|c|cccc|cccc|cccc}
\hline
\multicolumn{1}{c|}{\multirow{2}{*}{Model}} & \multirow{2}{*}{\begin{tabular}[c]{@{}c@{}}Training \\ Dataset\end{tabular}} & \multicolumn{4}{c|}{No Routine} & \multicolumn{4}{c|}{Routine w/ Branch} & \multicolumn{4}{c}{Routine w/o Branch} \\ \cline{3-14} 
\multicolumn{1}{c|}{} &  & Structural & Tool & Parameter & Overall & Structural & Tool & Parameter & Overall & Structural & Tool & Parameter & Overall \\ \hline
GPT-3.5-Turbo & \textbackslash{} & 99.8 & 26.0 & 95.3 & 22.1 & 99.8 & 53.8 & 98.2 & 52.7 & 99.9 & 59.7 & 98.1 & 58.5 \\
GPT-4-Turbo & \textbackslash{} & 100 & 44.6 & 97.5 & 43.5 & 100 & 99.1 & 99.8 & 99.0 & 100 & 98.7 & 99.6 & 98.3 \\
GPT-4o & \textbackslash{} & 100 & 42.2 & 97.3 & 41.1 & 100 & 96.3 & 99.9 & 96.3 & 100 & 97.0 & 99.9 & 97.0 \\
Claude-3.7-Sonnet & \textbackslash{} & 97.7 & 46.5 & 96.6 & 43.9 & 100 & 99.7 & 99.7 & 99.3 & 99.2 & 99.6 & 99.3 & 98.0 \\ \hline
Qwen2.5-7B & \textbackslash{} & 99.7 & 15.6 & 95.5 & 14.9 & 98.6 & 51.4 & 98.1 & 49.7 & 98.5 & 57.9 & 98.6 & 56.3 \\
Qwen2.5-14B & \textbackslash{} & 99.3 & 20.3 & 96.5 & 19.4 & 97.8 & 83.4 & 96.9 & 79.1 & 98.4 & 82.7 & 97.2 & 79.2 \\
Qwen3-8B & \textbackslash{} & 99.2 & 37.3 & 95.3 & 35.3 & 99.2 & 83.9 & 97.6 & 81.3 & 99.0 & 84.8 & 96.8 & 81.3 \\
Qwen3-14B & \textbackslash{} & 94.1 & 36.5 & 94.9 & 32.6 & 96.0 & 88.3 & 98.3 & 83.3 & 97.0 & 87.8 & 98.2 & 83.6 \\ \hline
Qwen2.5-7B & \textit{Common} & 93.6 & 25.1 & 95.2 & 22.4 & 97.6 & 89.0 & 98.6 & 85.7 & 96.9 & 90.7 & 98.7 & 86.8 \\
Qwen2.5-14B & \textit{Common} & 91.1 & 34.6 & 98.1 & 30.9 & 97.2 & 90.0 & 98.4 & 86.1 & 97.6 & 94.3 & 98.8 & 90.9 \\
Qwen3-8B & \textit{Common} & 95.3 & 26.3 & 93.4 & 23.4 & 96.3 & 88.3 & 97.2 & 82.8 & 97.4 & 94.4 & 97.2 & 89.3 \\
Qwen3-14B & \textit{Common} & 97.3 & 35.1 & 97.2 & 33.2 & 98.2 & 92.0 & 97.7 & 88.2 & 99.0 & 94.8 & 98.8 & 92.7 \\ \hline
Qwen2.5-7B & \textit{Scenario} & 99.7 & 88.8 & 99.1 & 87.8 & 99.7 & 94.1 & 99.7 & 93.5 & 99.7 & 95.2 & 99.5 & 94.4 \\
Qwen2.5-14B & \textit{Scenario} & 99.8 & 88.1 & 99.4 & 87.5 & 100 & 97.9 & 99.6 & 97.5 & 99.8 & 98.4 & 99.8 & 98.1 \\
Qwen3-8B & \textit{Scenario} & 100 & 89.4 & 99.4 & 88.9 & 99.8 & 94.4 & 99.3 & 93.6 & 99.9 & 96.9 & 99.3 & 96.2 \\
Qwen3-14B & \textit{Scenario} & 99.7 & 90.9 & 99.4 & 90.2 & 99.7 & 95.9 & 99.9 & 95.5 & 99.7 & 98.3 & 99.9 & 98.0 \\ \hline
\end{tabular}%
}
\end{table}

\subsubsection{Impact of Routine on LLM Tool Calling Performance}

Without Routine guidance, all baseline models performed poorly, with none exceeding 50\% overall accuracy. This indicates that even for top models, relying entirely on their autonomous planning ability for complex, multi-step tasks introduces significant uncertainty. \textbf{Tool selection errors were identified as the main cause of failure, accounting for over 85\% of all errors}. This finding reveals the huge challenge models face in accurately selecting from a large pool of available tools (over 25) and organizing them into an effective execution chain within a specialized domain.

The introduction of the Routine mechanism led to a substantial improvement in the models' end-to-end tool calling accuracy. In particular, GPT-4-Turbo's performance approached perfection, and the Qwen series models also demonstrated significant gains. This indicates that setting a well-defined Routine plan can significantly reduce the uncertainty in a model's execution process. The Routine mechanism can therefore effectively compensate for the planning deficiencies of smaller models, enabling them to achieve performance close to top models in specific scenarios. An analysis of the error distribution confirms that \textbf{the improvement in tool selection accuracy is the main driver of the overall accuracy increase}. This suggests that Routine effectively guides models to select the correct tools by decomposing tasks into clear, actionable steps.  Structural and parameter errors also saw concurrent improvements; although these errors were less frequent in baseline tests, they were further minimized under Routine guidance.

A comparison of the \textit{w/ Branch} and \textit{w/o Branch} Routine scenarios revealed that performance was generally higher in the \textit{w/o Branch} scenario. The performance difference was small for high-performing models. However, for models with average performance, the introduction of branches led to a more pronounced decline in accuracy. This suggests that the use of branching logic within Routines is most effective when built upon a model's already robust tool calling foundation to avoid performance degradation.

\subsubsection{Impact of Model Training}

The experimental results in Table \ref{tab:models-performance-on-routine} show that fine-tuning with the common Routine following dataset effectively improves the model's execution accuracy when a Routine is provided. Compared to their baselines, these models showed significant improvements across all metrics in the \textit{w/ Routine} scenarios. However, under the \textit{No Routine} condition that requires autonomous planning, the performance of these fine-tuned models declined, indicating a trade-off where their common problem-solving ability was reduced. This is because the strategy reinforces the model's role as a plan executor but does not enhance its core capability as an autonomous planner.

In contrast, fine-tuning via scenario-specific data distillation achieved a significant improvement in overall accuracy under the \textit{No Routine} condition, with performance exceeding original models even when they were guided by Routine. This indicates that for smaller models, using a teacher model guided by a Routine to distill data for training allows the procedural knowledge to be directly internalized within the student model, reducing the model's reliance on an explicit plan. Furthermore, when a Routine is provided to these already-specialized models, their accuracy is enhanced even further, approaching the level of GPT-4o and demonstrating highly stable execution. This confirms that injecting procedural knowledge both internally (via weights) and externally (via prompts) is an effective strategy for maximizing the model's stability in a target scenario.

\subsection{Ablation Study} 
Having established the effectiveness of Routine, we conducted a series of ablation studies to explore the impact of different Routine mechanism settings on overall accuracy. These studies focused on three main areas: the components of the Routine, the method of Routine annotation, and the quantity of Routines provided.

\subsubsection{Ablation on Routine Components}

We investigated the specific impact of a Routine's different internal components on the model's final execution accuracy, primarily testing the effects of tool specifications and detailed input/output parameter descriptions. We established three experimental conditions under the complex with branch scenario:

\begin{itemize}
\item \textbf{Baseline:} The model receives a complete natural language Routine with step descriptions and tool names, but without detailed parameter guidance.  

\item \textbf{With I/O Descriptions:} Building on the baseline, detailed descriptions of the input source and expected output are added to each step.

\item \textbf{Without Tool Name:} Building on the baseline, the direct instruction specifying which tool to select is removed from each step, requiring the model to infer the appropriate tool from the step's description.
\end{itemize}

Tests on representative models (Table \ref{tab:routine-components-ablation}) reveal the distinct impact of Routine components:

\begin{table}[htbp]
\centering
\caption{Overall Model Accuracy on Different Routine Components}
\label{tab:routine-components-ablation}
\begin{tabular}{l c c c}
\toprule
\textbf{Model} & \makecell{\textbf{Baseline Routine}\\} & \makecell{\textbf{Routine w/ I/O Params}\\} & \makecell{\textbf{Routine w/o Tools}\\} \\
\midrule
GPT-3.5-Turbo & 52.7 & 61.1 & 42.8 \\
GPT-4o & 96.3 & 97.5 & 96.7 \\
Qwen2.5-7b & 49.7 & 60.6 & 43.7 \\
Qwen2.5-14b & 79.1 & 84.2 & 69.5 \\
Qwen3-8b & 81.3 & 81.3 & 76.7 \\
Qwen3-14b & 83.3 & 81.5 & 71.9 \\
\bottomrule
\end{tabular}
\end{table}

Adding detailed parameter descriptions had a varied effect. For less capable models, this explicit guidance was highly effective, improving their contextual understanding and significantly reducing parameter errors. For high-performance models, the additional detail helped in making more robust judgments in edge cases, leading to incremental performance gains. Interestingly, the Qwen3 series models appeared largely insensitive to this addition, suggesting that for some architectures, the baseline Routine already contains enough context for parameter inference, and further verbosity may interfere model. Overall, \textbf{including I/O parameter descriptions in Routine is an effective strategy for improving an agent's stability and applicability}, as it offers essential context for moderately capable models and provides minor benefits even for leading models.  

The result confirms that \textbf{explicitly specifying the tool name within Routine is a core element for ensuring accurate execution}. When the tool name was removed, all models except the highly stable GPT-4o experienced a significant drop in accuracy (typically by 5\%-15\%). This demonstrates that providing the tool name transforms a difficult reasoning problem \textit{which tool to use} into a straightforward execution task \textit{use this tool}, thereby reducing the cognitive load required to understand the task logic. While GPT-4o demonstrated exceptional semantic understanding by inferring the correct tool from the step description alone, this was an outlier. For the majority of models, explicitly specifying the tool within the plan proved to be highly beneficial. 

Based on this analysis, we conclude that a well-designed Routine should function as a structured execution plan containing both explicit tool instructions and sufficient descriptions, such as I/O parameters. This ensures the agent can complete tasks with maximum stability and accuracy, regardless of the underlying driving model's capabilities.

\subsubsection{Impact of AI-Optimization Mechanism}

\begin{wrapfigure}{r}{0.56\linewidth} 
    \centering
    \includegraphics[width=1\linewidth]{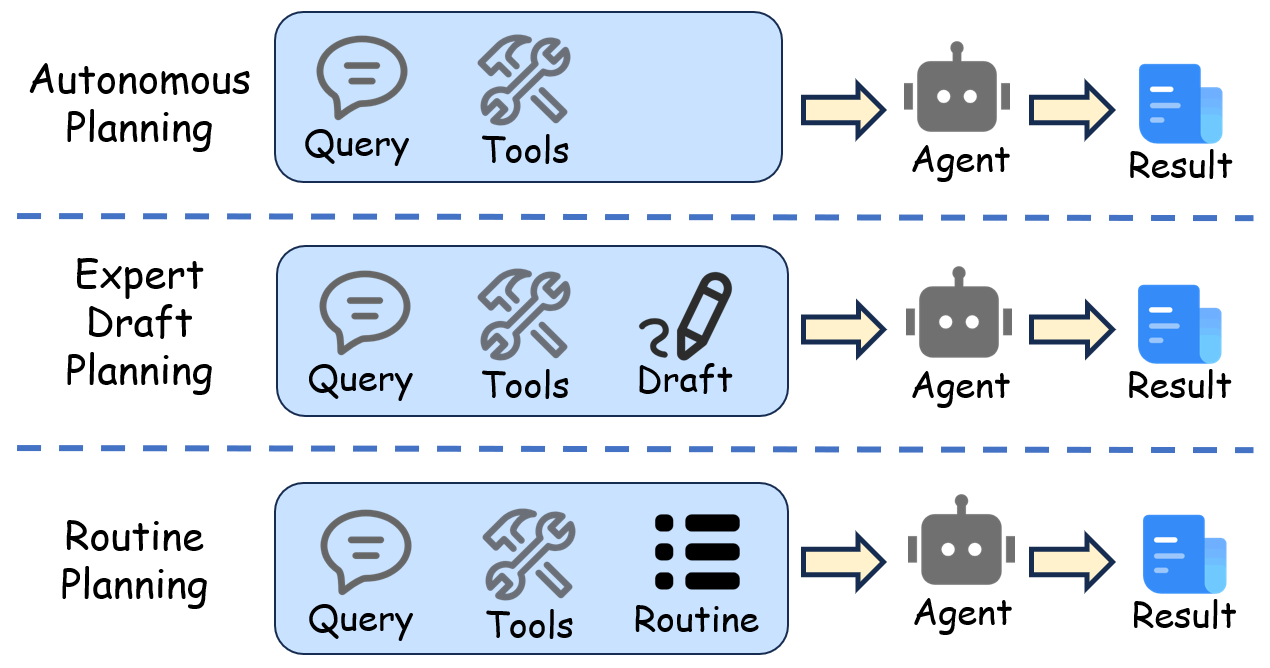}
    \vspace{-0.5em}
    \caption{Different ways of Agent Planning}
    \vspace{-0.5em}
    \label{fig:threeLine}
\end{wrapfigure}

In practice, there are different ways of agent planning, as shown in Figure \ref{fig:threeLine}. Using a manual Routine brings high robustness, but its annotation is costly and unscalable. Therefore, we explored AI-driven optimization as a practical alternative.  This ablation study compared the Routine of varying refinement levels:

\begin{itemize}
\item \textbf{User-Drafted Routine:} An initial, incomplete, and unstructured natural language prompt from a user,  containing only the basic sequence of steps. This is the starting point for evaluating the effectiveness of optimization.

\item \textbf{AI-Optimized Routine:} The user draft is automatically corrected, completed, and refined by GPT-4o and sets a corresponding tool selection for each step, forming a more logical and complete natural language Routine. 

\item \textbf{Manually Annotated Routine:} A complete natural language Routine with branches, meticulously annotated and calibrated by domain experts.
\item \textbf{Value of the Initial Draft:} Even a low-quality user draft improved model execution accuracy. This demonstrates that even a rough plan provides crucial guidance for models and outperforms fully autonomous tool calling in specialized domains.
\end{itemize}

Tests on representative models (Table \ref{tab:routine-generation-ablation}) reveal the distinct impact of different Routine generation methods:

\begin{table}[htbp]
\centering
\caption{Overall model accuracy on different generation methods of Routine}
\label{tab:routine-generation-ablation}
\begin{tabular}{l c c c}
\toprule
\textbf{Model} & \makecell{\textbf{User Draft}\\} & \makecell{\textbf{AI Optimization}\\} & \makecell{\textbf{Human Annotation}\\} \\
\midrule
GPT-3.5-Turbo & 42.6 & 52.9 & 52.70 \\
GPT-4o & 71.2 & 90.9 & 96.3 \\
Qwen2.5-7b & 46.6 & 50.4 & 49.7 \\
Qwen2.5-14b & 61.7 & 82.3 & 79.1 \\
Qwen3-8b & 70.4 & 73.8 & 81.3 \\
Qwen3-14b & 70.9 & 76.7 & 83.3 \\
\bottomrule
\end{tabular}
\end{table}

The AI optimization step brought significant and universal performance improvements. The performance of GPT-4o, bridged most of the gap between basic usability and high reliability. The Qwen3 series models have also achieved stable improvements. Some models even achieved higher accuracy with AI-optimized Routine than with the manually annotated baseline, possibly because the AI-optimized format was more easily understood by those models. This shows that using AI to refine a user-drafted Routine is an efficient and viable pathway for enterprise scenarios.

However, the manually annotated baseline still produced the highest accuracy for high-performance models. The details provided by human experts remain crucial, as these models are capable enough to benefit from these subtle details. Therefore, in real-world applications, a final review of the AI-optimized Routine by domain experts is still recommended.

\subsubsection{Ablation on Routine Quantity}

An agent system often needs to handle multiple sub-scenarios, requiring a library of Routines. Loading all Routines into the system prompt would consume a large context window. To address this, the system can implement a procedure memory recall mechanism in the memory module as mentioned in Section 3.4. However, recalling a single Routine can easily cause precision issues, while recalling multiple Routines can introduce noise from interfering Routines that are similar but not perfectly applicable. This study tests the model's stability when faced with multiple candidate Routines. The setup is as follows:
 
\begin{itemize}
\item \textbf{Baseline (1 Routine):} The model receives only one correct Routine.

\item  \textbf{Multi-Routine Interference Scenario:} The model is provided with 2, 3, or 5 Routines in total, each labeled with its name and function. Only one is applicable; the rest are interferences that are not related to this sub-scenario. The order of Routines is shuffled to prevent positional bias.
\end{itemize}

The results of the tests are shown in (Table \ref{tab:routine-number-ablation}), which reflects the impact of multiple numbers of Routines:

\begin{table}[htbp]
\centering
\caption{Overall model accuracy on multiple numbers of Routines}
\label{tab:routine-number-ablation}
\begin{tabular}{l c c c c}
\toprule
\textbf{Model} & \makecell{\textbf{1 Routine(Baseline)}\\} & \makecell{\textbf{2 Routines}\\} & \makecell{\textbf{3 Routines}\\} & \makecell{\textbf{5 Routines}\\} \\
\midrule
GPT-3.5-Turbo & 52.7 & 53.6 & 54.7 & 60.0 \\
GPT-4o & 96.3 & 76.6 & 80.1 & 88.6 \\
Qwen2.5-7b & 49.7 & 45.1 & 45.0 & 54.4 \\
Qwen2.5-14b & 79.1 & 55.8 & 60.7 & 71.3 \\
Qwen3-8b & 81.3 & 66.0 & 66.9 & 72.2 \\
Qwen3-14b & 83.3 & 63.2 & 67.2 & 79.9 \\
\bottomrule
\end{tabular}
\end{table}

The results (Table \ref{tab:routine-number-ablation}) show that providing a single correct Routine is most effective for high-performing models. Their accuracy dropped significantly when even one distractor was introduced. However, as the number of recalled Routines increased, the accuracy began to recover. We hypothesize a shift in the model's behavior: when faced with a small number of conflicting Routines, it may attempt to combine their steps, leading to errors. As the number of Routines increases, it may switch to a selection mechanism (similar to tool selection), identifying and executing the most relevant Routine, thus improving performance.

Performance fluctuations are greater for smaller models, sometimes even increasing with more numbers of Routines. We believe that this is caused by overlapping sub-steps of some Routines. As more Routines are recalled, these common sub-steps repeatedly appear, which may cause the model to allocate extra attention to them. This repetition improves the execution accuracy of these specific sub-steps for less capable models, resulting in an increase in the overall scenario accuracy.

The experiment indicates that an agent's memory module must be carefully designed and optimized to achieve high precision and recall, aiming to provide only one single and most relevant Routine to the execution model to enhance system efficiency and stability.
\section{Limitation and Future Work}
The Routine mechanism we proposed has demonstrated its effectiveness in enhancing an agent system's ability to solve tasks via multi-step tool invocation within specific scenarios. However, current planning models primarily rely on drafts provided by domain experts to generate Routine flows, while execution models are mostly adapted through instruction fine-tuning via knowledge distillation. This reliance limits the system's generalization capability when new tools are introduced or workflow changes occur within enterprise environments, leaving room for improvement in agent autonomy and adaptability.

To address this challenge, incorporating RL-based agent frameworks into the workflow, including mechanisms for data distillation and reward modeling might be a possible solution. This approach aims to improve the Routine generation capability of the planning model as well as the tool invocation capability of the execution model. The combination of instruction fine-tuning for cold start and reinforcement learning has shown promising potential in improving both generalization and adaptability in scenario-based tasks, and may emerge as a future paradigm for training agent-based language models~\cite{huan2025does}.

Furthermore, we aim to explore a multi-agent framework centered around the Routine mechanism, in which a high-level agent coordinates multiple specialized agents through a set of structured Routine flows and centralized interaction protocols. We hypothesize that this hierarchical interaction scheme can effectively reduce the complexity and length of individual Routine plans, thereby enabling more stable and intelligent execution of enterprise workflows. Through this line of research, our goal is to develop more intelligent, robust, and adaptive LLM-based agents capable of leveraging tools efficiently to solve complex user problems in dynamic enterprise environments.
\section{Conclusion}

In this paper, we design Routine, a structured and comprehensive planning framework for guiding multi-step tool execution in agent systems. Using Routine, we investigate how well-defined plans improve the accuracy of the model's multi-step execution. We also synthesize a training dataset to enhance Routine distillation capabilities and generate domain-specific, multi-step tool-calling datasets via Routine-based distillation. Our experimental results demonstrate that Routines significantly improve the execution model's accuracy, improving the performance of the Qwen3-14b model by approximately 50\%, from 32.6\% to 83.3\%. This accuracy is further increased after fine-tuning the model for Routine following. Furthermore, by using Routines to distill training datasets for scenario-specific fine-tuning, the performance of Qwen3-14b has been improved to 95.5\%, which is comparable to GPT-4o, enabling smaller models to achieve high and stable accuracy in enterprise scenarios. In conclusion, the Routine mechanism significantly improves the adaptability of agent systems to enterprise scenarios. It allows AI to assist more effectively in the execution of enterprise processes, thereby realizing the technical vision of \textbf{AI for Process}.\\

\bibliographystyle{unsrt}
\bibliography{References}

\newpage
\appendix
\section{Appendix: Prompt Template} \label{sec:apx_a}
\subsection{Prompt for Structured Routine Generation}

\lstset{
  commentstyle=\color{black},    
  keywordstyle=\color{black},       
  stringstyle=\color{black},     
  columns=fullflexible,
  breakindent=20pt,
}

\begin{lstlisting}
prompt = f"""You are a Routine workflow writer for a company. You can write the operation step flow based on the process information provided by the user and the available tools.
The steps are written in structured json and lists. Write the flow in the following way:

[{"step": "1", "name": "xxxxx", "description": "xxxxxxxxxxxx", "tool": "tool_X", "type": "node"},
{"step": "2", "name": "xxxxx", "description": "xxxxxxxxxxxx", "tool": "tool_Y", "type": "node"}]

The format is a json list. Each step contains the step number, step name, step action description, step input, step output, step tool, and node type.
The input and output of the step do not have to be very specific. Use natural language to write the possible input and output according to the tool. Only one tool is used for each step.
When you may encounter branch condition judgment in a certain step, express it in the following way and indicate under what conditions to enter a branch, what tool to use;

{"step": "x", "name": "xxxxx", "type": "branch"},
    {"step": "x-1_1", "name": "xx", "description": "xxxx", "tool": "tool_X1", "type": "branchnode"}, 
    {"step": "x-2_1", "name": "xx", "description": "xxxx", ""tool": "tool_X2", "type": "branchnode"},
{"step": "y", "name": "xxxxx", "description": "xxxxxx", "tool": "tool_Y", "type": "node"}

If the next branch step involves multiple steps, you can open a new branch workflow, for example:
{"step": "x-n_1", "name": "xx", "description": "xxxx", "tool": "tool_X", "type": "branchnode"}, 
{"step": "x-n_2", "name": "xx", "description": "xxxx", "tool": "tool_Y", "type": "branchnode"}

Regarding the writing of step numbers, x-n_i represents the i-th step in the n-th branch of the main line step x;
Please pay attention to the description in the tool and the parameters that need to be filled in, which need to be fed back in the input of each step;
Pay attention to the branch judgment in the process information, and do not write multiple possibilities of branch conditions in the steps of the same line;
When a step is completed and the workflow needs to be ended, please change the node type of the step to "finish", set "type": "finish"; For example: 

{"step": "x", "name": "xxxxx", "description": "xxx", "tool": "tool_X", "type": "finish"}

Note: Each workflow step must use a tool provided in the tool list, or perform branch condition judgment. There will be no "no tool needed", "no tool used", or use of non-existent tools. Each step only uses one tool. 
The following is the process information provided by this user: {routine_draft};
In the tool list, these tools are available: {tool_list};
Now please convert it into a structured Routine workflow. Do not output other prefixes, suffixes, or meaningless information, and please output in Chinese."""
\end{lstlisting}

With the above prompt template, LLMs can convert the user planning draft into a routine. The output Routine is in a standardized JSON format and can be further converted into a natural language Routine as shown in \ref{sec:apx_NL_Routine}; \\

\subsection{Routine's System Prompt Template for Tool Calling}
\begin{lstlisting}
prompt = f"""<|im_start|>system
You are a digital HR of a company, and you can use multiple different types of tools to query relevant data of user questions and answer user questions;
You have a scenario workflow operation step called Routine. You need to select a tool to call based on the steps in the Routine and the completed historical steps;
You have completed tool calls for similar scenarios before and have a memory of tool calls for similar scenarios. Now you can imitate the previous tool calls to select the tool you need to call now based on the historical dialogue information;
Please strictly imitate the tool call instruction steps in the Routine, do not add tool call instructions that have not appeared in the Routine, and only output one tool call at a time;
In the case of branches, please judge the branches of subsequent steps according to the conditions of each branch;
Note: When temporary variable memory_xxx appears in the result returned by the tool result, it means that the value of the variable xxx is too long and is stored in the temporary variable memory. Try to fill in temporary variable memory_xxx instead of the actual value in the tool call parameters;
The user id of the current question is USER_ID;

# Routine
To solve user questions, you need to refer to the following routines, select the tool you need to call, and make function calls based on the current progress and chat history. 
Please strictly follow the routines, do not skip any steps, and only output one function call at a time.
The <routines></routines> XML tag provides you with the routine signatures of the workflow operation steps:

<routines>
{ROUTINES_DESCRIPTION}
</routines>

# Variables
To ensure smooth information flow between each step, the system puts the following temporary variables into the <variables></variables> XML tag to record the intermediate results:

<variables>
{VARIABLES_MEMORY}
</variables>

# Tools
You may call one or more functions to assist with the user query.
You are provided with function signatures within <tools></tools> XML tags:

<tools>
TOOLS
</tools>

For each function call, return a json object with function name and arguments within <tool_call></tool_call> XML tags:
<tool_call>
{"name": <function-name>, "arguments": <args-json-object>}
</tool_call><|im_end|>
"""
\end{lstlisting}

\newpage
\section{Appendix: Routine and Dataset Example} \label{sec:apx_b}

\subsection{Routine for Tool Calling in Natural Language } \label{sec:apx_NL_Routine}
\begin{lstlisting}
"""
<routines>
Step 1. Get announcements: Download the latest employee handbook file from the company's internal system, use the fetch_latest_announcements tool;

Step 2. Download handbook: Obtain the company's most recent official announcement for consistency check with the employee handbook, use the download_file tool;

Step 3. Read PDF content: Use text parsing tools to extract all text content from the employee handbook PDF file, use the read_pdf tool;

Step 4. Compare text differences: Compare the relevant content in the employee handbook with the company's latest announcement word by word, using the compare_texts tool, and end the workflow;
</routines>
"""
\end{lstlisting}

\subsection{Structured Routine for Tool Calling in JSON Format}
\begin{lstlisting}
[{
"step": "1"
"name": "Get announcements"
"description": "Download the latest employee handbook file from the company's internal system"
"tool": "fetch_latest_announcements"
"type": "node"
}, 
{
"step": "2"
"name": "Download handbook"
"description": "Obtain the company's most recent official announcement for consistency check with the employee handbook"
"tool": "download_file"
"type": "node"
}, 
{
"step": "3"
"name": "Read PDF content"
"description": "se text parsing tools to extract all text content from the employee handbook PDF file"
"tool": "read_pdf" 
"type": "node"
},
{
"step": "4"
"name": "Compare text differences"
"description": "Compare the relevant content in the employee handbook with the company's latest announcement word by word"
"tool": "compare_texts" 
"type": "finish"
}]

\end{lstlisting}

\newpage

\subsection{A Multi-Step Tool Calls Dataset Example with Routines in ShareGPT Format}

\begin{lstlisting}
{
"conversations": [
  {
    "from": "human",
    "value": "Hi, I need to download the \"Employee Handbook 2023\" PDF from our company's internal HR portal. Once I have it, I need to check if there are any updates in the handbook based on the latest announcements from the HR portal. Can you assist me with this?"
  },
  {
    "from": "function_call",
    "value": "{\"name\": \"fetch_latest_announcements\", \"arguments\": {\"portal_url\": \"https://company-hr-portal.com/announcements\"}}"
  },
  {
    "from": "observation",
    "value": "{\"announcements\": [\"New company policies will be effective from next month.\", \"Annual company picnic scheduled for September 15th.\", \"Employee of the month for August: John Doe.\", \"Reminder: Submit your performance reviews by the end of this week.\"]}"
  },
  {
    "from": "function_call",
    "value": "{\"name\": \"download_file\", \"arguments\": {\"url\": \"https://company-hr-portal.com/employee-handbook-2023.pdf\", \"destination_path\": \"/local/path/employee-handbook-2023.pdf\"}}"
  },
  {
    "from": "observation",
    "value": "{\"file_path\": \"/local/path/employee-handbook-2023.pdf\"}"
  },
  {
    "from": "function_call",
    "value": "{\"name\": \"read_pdf\", \"arguments\": {\"file_path\": \"/local/path/employee-handbook-2023.pdf\"}}"
  },
  {
    "from": "observation",
    "value": "{\"content\": \"Welcome to the 2023 Employee Handbook. This document outlines policies, procedures, and benefits available to employees. Chapter 1: Introduction. ...\"}"
  },
  {
    "from": "function_call",
    "value": "{\"name\": \"compare_texts\", \"arguments\": {\"text1\": \"Welcome to the 2023 Employee Handbook. This document outlines policies, procedures, and benefits available to employees. Chapter 1: Introduction. ...\", \"text2\": \"New company policies will be effective from next month.\\nAnnual company picnic scheduled for September 15th.\\nEmployee of the month for August: John Doe.\\nReminder: Submit your performance reviews by the end of this week.\"}}"
  }
],
"system": "
    You are an expert at calling functions (tools), and can accurately select and call the   appropriate tool from the available tool set according to the user's task to answer the user's questions; You have a scenario workflow operation step called Routine. You need to select a tool to call based on the steps in the Routine and the completed historical steps; You have completed tool calls for similar scenarios before and have a memory of tool calls for similar scenarios. Now you can imitate the previous tool calls to select the tool you need to call now based on the historical dialogue information; Please strictly imitate the tool call instruction steps in the Routine, do not add tool call instructions that have not appeared in the Routine, and only output one tool call at a time; In the case of branches, please judge the branches of subsequent steps according to the conditions of each branch;

    Note: When temporary variable memory_xxx appears in the result returned by the tool result, it means that the value of the variable xxx is too long and is stored in the temporary variable memory. Try to fill in temporary variable memory_xxx instead of the actual value in the tool call parameters;

    # Routine
    To solve user questions, you need to refer to the following routines, select the tool you need to call, and make function calls based on the current progress and chat history. Please strictly follow the routines, do not skip any steps, and only output one function call at a time.
    The <routines></routines> XML tag provides you with the routine signatures of the workflow operation steps:
    
    <routines>
    Step 1. Get announcements: Download the latest employee handbook file from the company's internal system, use the fetch_latest_announcements tool;
    
    Step 2. Download handbook: Obtain the company's most recent official announcement for consistency check with the employee handbook, use the download_file tool;
    
    Step 3. Read PDF content: Use text parsing tools to extract all text content from the employee handbook PDF file, use the read_pdf tool;
    
    Step 4. Compare text differences: Compare the relevant content in the employee handbook with the company's latest announcement word by word, using the compare_texts tool, and end the workflow;
    </routines>

    # Variables
    To ensure smooth information flow between each step, the system uses the following temporary variables to record intermediate results:
    
    <variables>
    </variables>
"
"tools": "[{\"name\": \"compare_texts\", \"description\": \"Compares two sets of texts to find differences or updates.\", \"parameters\": {\"type\": \"object\", \"properties\": {\"text1\": {\"type\": \"string\", \"description\": \"The first text to compare (e.g., the existing handbook content).\"}, \"text2\": {\"type\": \"string\", \"description\": \"The second text to compare (e.g., the latest announcements).\"}}, \"required\": [\"text1\", \"text2\"]}}, {\"name\": \"read_pdf\", \"description\": \"Reads the text content from a PDF file.\", \"parameters\": {\"type\": \"object\", \"properties\": {\"file_path\": {\"type\": \"string\", \"description\": \"The path of the PDF file to read.\"}}, \"required\": [\"file_path\"]}}, {\"name\": \"download_file\", \"description\": \"Downloads a file from a given URL to a specified local path.\", \"parameters\": {\"type\": \"object\", \"properties\": {\"url\": {\"type\": \"string\", \"description\": \"The URL of the file to download.\"}, \"destination_path\": {\"type\": \"string\", \"description\": \"The local path to save the downloaded file.\"}}, \"required\": [\"url\", \"destination_path\"]}}, {\"name\": \"fetch_latest_announcements\", \"description\": \"Fetches the latest announcements or updates from the given portal URL.\", \"parameters\": {\"type\": \"object\", \"properties\": {\"portal_url\": {\"type\": \"string\", \"description\": \"The URL of the portal to fetch announcements from.\"}}, \"required\": [\"portal_url\"]}}]"
}
\end{lstlisting}

\end{document}